\pdfoutput=1
\documentclass[sigconf,nonacm]{acmart}

\usepackage{array}
\usepackage{booktabs} 
\usepackage{epstopdf}
\usepackage{multirow}
\usepackage{balance}
\usepackage{amsfonts}
\usepackage{amsmath}
\usepackage{amssymb}
\usepackage{graphicx}
\usepackage{subfigure}
\usepackage[noend,ruled]{algorithm2e}
\usepackage{microtype}
\usepackage{url}
\usepackage{enumitem}
\usepackage{tabularx}
\usepackage{caption} 
\usepackage{commath}  
\usepackage{makecell}  

\copyrightyear{2021}
\acmYear{2021}
\setcopyright{iw3c2w3}
\acmConference[WWW '21]{Proceedings of the Web Conference 2021}{April 19--23, 2021}{Ljubljana, Slovenia}
\acmBooktitle{Proceedings of the Web Conference 2021 (WWW '21), April 19--23, 2021,
Ljubljana, Slovenia}

\acmPrice{}
\acmDOI{10.1145/3442381.3450114}
\acmISBN{978-1-4503-8312-7/21/04}

\def\ddefloop#1{\ifx\ddefloop#1\else\ddef{#1}\expandafter\ddefloop\fi}
\def\ddef#1{\expandafter\def\csname bb#1\endcsname{\ensuremath{\mathbb{#1}}}}
\ddefloop ABCDEFGHIJKLMNOPQRSTUVWXYZ\ddefloop
\def\ddef#1{\expandafter\def\csname bf#1\endcsname{\ensuremath{\mathbf{#1}}}}
\ddefloop ABCDEFGHIJKLMNOPQRSTUVWXYZabcdefghijklmnopqrstuvwxyz\ddefloop
\def\ddef#1{\expandafter\def\csname c#1\endcsname{\ensuremath{\mathcal{#1}}}}
\ddefloop ABCDEFGHIJKLMNOPQRSTUVWXYZ\ddefloop
\def\ddef#1{\expandafter\def\csname v#1\endcsname{\ensuremath{\boldsymbol{#1}}}}
\ddefloop ABCDEFGHIJKLMNOPQRSTUVWXYZabcdefghijklmnopqrstuvwxyz\ddefloop
\def\ddef#1{\expandafter\def\csname v#1\endcsname{\ensuremath{\boldsymbol{\csname #1\endcsname}}}}
\ddefloop {alpha}{beta}{gamma}{delta}{epsilon}{varepsilon}{zeta}{eta}{theta}{vartheta}{iota}{kappa}{lambda}{mu}{nu}{xi}{pi}{varpi}{rho}{varrho}{sigma}{varsigma}{tau}{upsilon}{phi}{varphi}{chi}{psi}{omega}{Gamma}{Delta}{Theta}{Lambda}{Xi}{Pi}{Sigma}{varSigma}{Upsilon}{Phi}{Psi}{Omega}{ell}\ddefloop

\def\R{\mathbb{R}}

\newtheorem{thm:def}{Definition}
\newtheorem{thm:eg}{Example}
\newtheorem{thm:lem}{Lemma}
\newtheorem{thm:obs}{Observation}
\newtheorem{thm:req}{Requirement}

\newcommand{\nop}[1]{}

\newcommand{\method}[1]{\mbox{\sf #1}\xspace}
\newcommand{\our}{\method{LTRN}}

\DeclareMathOperator*{\argmax}{argmax}

\DeclareMathAlphabet{\mathbbold}{U}{bbold}{m}{n}

\newcommand{\smallsection}[1]{\vspace{1mm}\noindent\textbf{#1.}}    

\begin{document}

\title{Minimally-Supervised Structure-Rich Text Categorization \\ via Learning on Text-Rich Networks}
\author{Xinyang Zhang$^{1}$, Chenwei Zhang$^{2}$, Luna Xin Dong$^{2}$, Jingbo Shang$^{3}$, Jiawei Han$^{1}$}
\affiliation{
    \institution{$^1$University of Illinois at Urbana-Champaign}
    \country{}
}
\affiliation{
    \institution{$^2$Amazon.com, Inc. \quad $^3$University of California San Diego}
    \country{}
}
\affiliation{
    $^1$\{xz43,hanj\}@illinois.edu \quad $^2$\{cwzhang, lunadong\}@amazon.com \quad $^3$jshang@ucsd.edu
    \country{}
}

\renewcommand{\shortauthors}{Xinyang Zhang et al.}
\renewcommand{\shorttitle}{Minimally-Supervised Structure-Rich Text Categorization via Learning on Text-Rich Networks}


\begin{abstract}

Text categorization is an essential task in Web content analysis.
Considering the ever-evolving Web data and new emerging categories, instead of the laborious supervised setting, in this paper, we focus on the minimally-supervised setting
that aims to categorize documents effectively, with a couple of seed documents annotated per category.
We recognize that texts collected from the Web are often structure-rich, i.e., accompanied by various metadata.
One can easily organize the corpus into a text-rich network, joining raw text documents with document attributes, high-quality phrases, label surface names as nodes, and their associations as edges.
Such a network provides a holistic view of the corpus' heterogeneous data sources
and enables a joint optimization for network-based analysis and deep textual model training.
We therefore propose a novel framework for minimally supervised categorization by learning from the text-rich network.
Specifically, we jointly train two modules with different inductive biases -- a text analysis module for text understanding and a network learning module for class-discriminative, scalable network learning.
Each module generates pseudo training labels from the unlabeled document set, and both modules mutually enhance each other by co-training using pooled pseudo labels.
We test our model on two real-world datasets.
On the challenging e-commerce product categorization dataset with 683 categories, our experiments show that given only three seed documents per category, our framework can achieve an accuracy of about 92\%, significantly outperforming all compared methods; our accuracy is only less than 2\% away from the supervised BERT model trained on about 50K labeled documents.

\end{abstract}

\maketitle


\section{Introduction}

\begin{figure}[t]
    \centering
    \includegraphics[width=\linewidth]{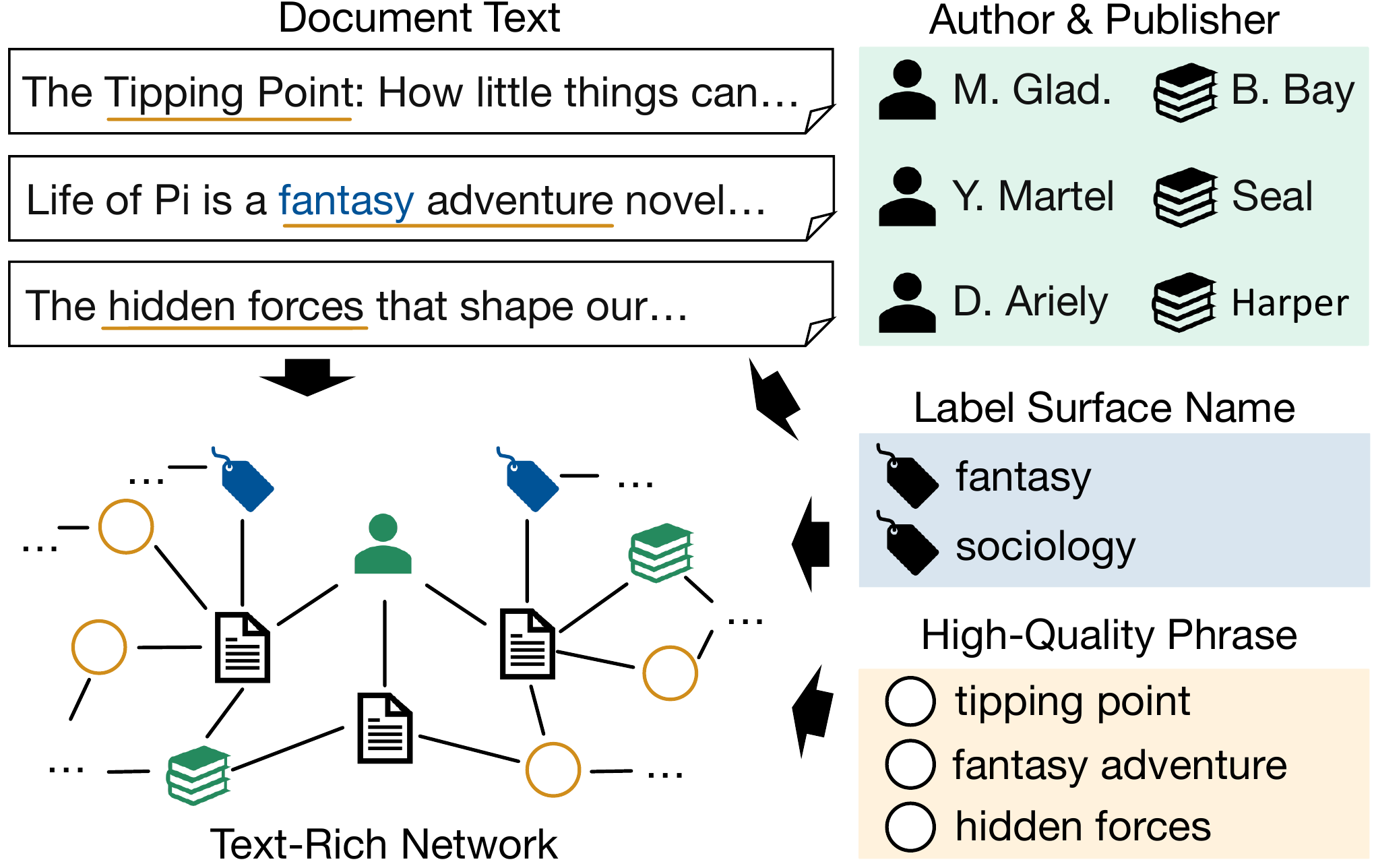}
    \vspace{-3mm}
    \caption{An illustrative example of a text-rich network constructed from e-books. It integrates the book descriptions (i.e., raw texts) and structure-rich metadata (e.g., author \& publisher, high-quality phrases, and label surface names) together and provides a holistic view.
    }
    \vspace{-5mm}
    \label{fig:text-rich}
\end{figure}

Text categorization is a fundamental task in organizing and understanding content gathered from the Web.
The Web's ever-evolving nature constantly brings in emerging categories to real-world text categorization, leading to hundreds of or even more categories. 
For example, e-commerce product categorization has tens of thousands of categories~\cite{sun2014product, chen2019product, song2020product}; classifying events from social media have hundreds of event types in commonly used event databases~\cite{leetaru2013gdelt}.
Therefore, instead of the laborious supervised setting, we set our focus on the \textit{minimally-supervised} setting---the user only provides a handful of examples per category (in our experiments, no more than 3 per category).
Our goal is to leverage these seed examples as well as unlabeled examples that are ubiquitously available to build an accurate categorization model.

In this paper, we provide a \textit{text-rich network} prospective for minimally-supervised categorization of \textit{structure-rich} text, i.e., text accompanied with metadata.
We argue that interconnecting documents and metadata together is beneficial in a minimally supervised setting.
Consider an example of book classification.
Given only a book intro, such as the one to the upper left in Figure~\ref{fig:text-rich}, it may be hard to tell the book's genre.
While if we look at its author and publisher and find more similar ``sociology'' books, we may have a better chance to put it into the correct category.
As illustrated in Figure~\ref{fig:text-rich}, we organize a structure-rich corpus into a text-rich network, integrating textual documents and various types of metadata into a single, unified framework.
It explicitly models their relations, offering a complementary view to raw text in the corpus, and enables network-based analysis.

We highlight two major challenges in modeling text-rich networks: (1) combining text and network for minimal supervision and (2) handling ``mixed signal'' from the text-rich network structure.
First, combining text and network structure is non-trivial, especially in the context of minimal supervision.
A desirable framework should take advantage of both modern natural language processing models for raw text understanding and graph learning models to propagate information on the network effectively.
However, trivially gluing two deep models from both data sources together may result in an over-complicated model that is infeasible computationally and prone to over-fitting on a small number of human labels.
Second, the network is by no means entirely constructed in a \textit{class-discriminative} way.
On the one hand, the network exhibits weak homophily, i.e., nodes sharing links or similar neighbors do not necessarily belong to the same category.
The problem becomes more evident as the number of categories becomes larger, and the category semantics become more similar.
As an example, 
when categorizing e-commerce products in different categories, ``screw-driver'' and ``wrench'' products can easily be confused because they usually share similar brands, manufacturers, and have similar wording in their product descriptions, making them close in the text-rich network.
On the other hand, the linkage in the network is far from perfect.
Continuing the product categorization example, products will be connected to nodes that are not label-indicative, such as the connection from a product to a phrasal node ``high-quality''.
The text-rich network's nature calls for a model that can identify label-discriminative features from the network.

While most existing methods~\cite{xie2019uda, inui2019eda, meng2018westclass, li2019learn-self} on minimally supervised text categorization employ text data only, pioneer studies on text-rich network either assume a relatively clean network structure for learning~\cite{tu2017cane, shen2018align}, or rely on additional human effort, such as user-given network motif patterns, to filter out uninteresting nodes~\cite{shang2020nettaxo, mekala2020meta}.
As a result, they do not address the aforementioned challenges in text-rich network modeling.

\begin{figure*}[t]
    \centering
    \includegraphics[width=\textwidth]{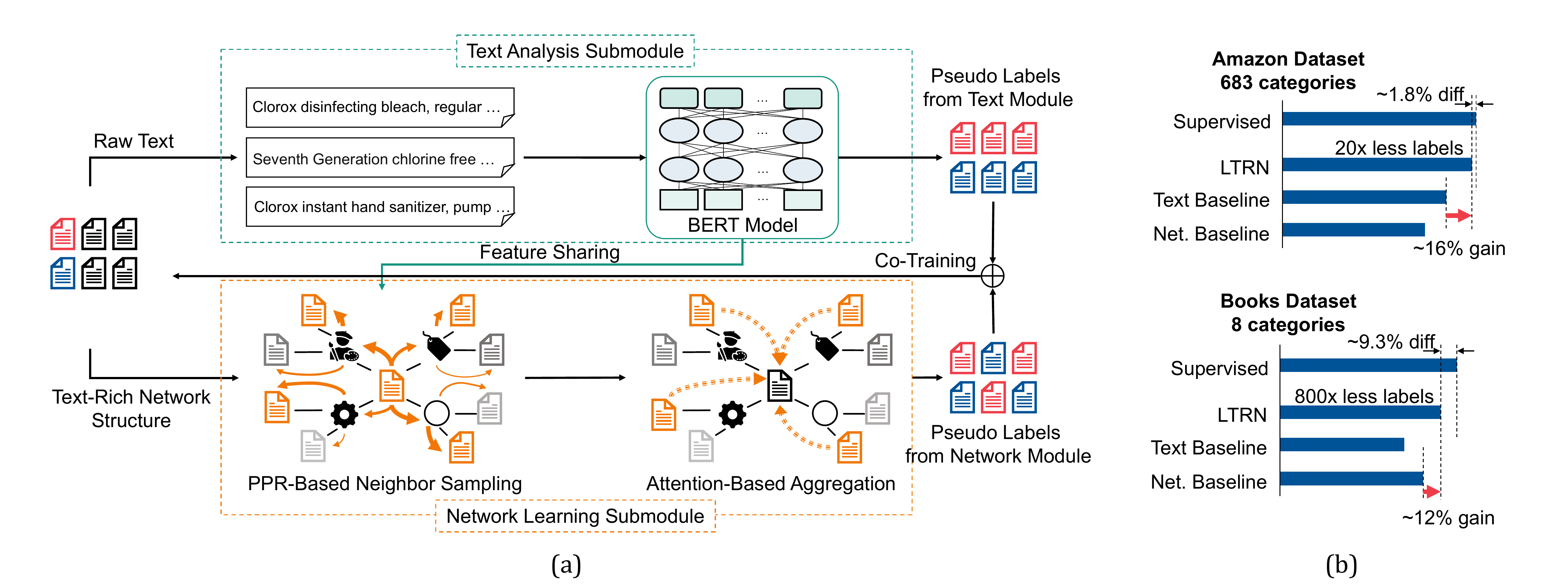}
    \vspace{-3mm}
    \caption{(a) An overview of \our. It jointly trains a text analysis module for raw text embedding and classification, and a scalable, noise-resilient network learning module for prediction on network structure. Both modules generate pseudo labels to extend minimal supervision, and are co-trained with feature sharing mechanism. (b) On two real-world datasets, with only 3 labeled seed examples per category, \our achieves significant gains over baselines and rivals the supervised 
   method.}
  \vspace{-3mm}
    \label{fig:framework}
\end{figure*}

To this end, we propose \our, a novel framework for minimally supervised text categorization by \textbf{L}earning on \textbf{T}ext-\textbf{R}ich \textbf{N}etworks.
As shown in Figure~\ref{fig:framework}, \our has two data-driven modules \textit{in parallel} -- a text analysis module for raw text embedding and classification, and a network learning module for semantic-aware propagation of categorical information on the network.
The two modules, one powered by BERT~\cite{devlin2019bert} and the other powered by a graph neural network (GNN), are designed to have very different \textit{inductive biases}.
The text module models raw text but is unaware of the network structure, while the network module effectively aggregates information from the network but relies on vectorized node features.
This offers us an opportunity to let each module learn from the other.
We leverage \textit{co-training} and \textit{feature sharing} to train both modules jointly.
In each iteration of our framework, we let both modules assign pseudo labels to the unlabeled documents, and employ co-training to pool two diverse sets of high confident predictions together, forming a pseudo labeled training set for the next algorithm iteration.
In this fashion, both modules can mutually enhance each other through the other model's predictions.
As the GNN requires fixed, vectorized features as input, at the end of each iteration, we share the document embedding from BERT to the GNN model, ensuring the GNN model gets the up-to-date document features.
After the framework is fully trained, we use the BERT model as our final categorization model.
The BERT model does not hinge on the network structure, thus is more flexible in both transductive and inductive setting when categorizing incoming new documents.

The specific design of each module in our framework is tailored to its data source.
The BERT model~\cite{devlin2019bert} in the text module leverages sentence-level semantics for text-based classification directly, and also provides meaningful document embedding for the network module.
The GNN model in the network module adopts a novel architecture with two proposed mechanisms to capture class-discriminative signals in the text-rich network.
First, a personalized PageRank (PPR) based neighborhood sampling method picks the most relevant neighbors for each document node.
Then an attention-based aggregation method rolls out unhelpful neighbors and further narrows down label-indicative neighbors.
Moreover, the model is scalable, making it suitable for gigantic real-world datasets.
Our network module design differs from the commonly used neighborhood sampling GNNs~\cite{hamilton2017graphsage, chen2018fastgcn} that do not model label-discriminativeness, and sets us apart from the popular Graph Attention Network~\cite{velickovic2018gat} which applies attention to the full neighborhood.

In summary, our main contributions are as follows:
\begin{itemize}[leftmargin=*,nosep]
    \item We propose a joint learning framework on the text-rich network for minimally-supervised text categorization.
    The framework has two data-driven modules with different inductive biases, a text analysis module, and a network learning module.
    We leverage co-training and feature sharing to train both modules jointly.
	\item We propose a novel GNN architecture for text-rich network modeling.
	The proposed model adopts personalized PageRank-based neighborhood sampling and attentive aggregation.
	The model successfully handles noise on the text-rich network and scales up to large datasets.
	\item We conduct experiments on two real-world datasets and achieve significant gain over various competitive baselines. On the challenging product categorization dataset with $\sim$700 categories, our model obtains an F-measure of over $90\%$ with only 3 seed per category.
	Comparing with a supervised BERT model, our model saves training labels by $20\times$ with only $<2\%$ sacrifice on F-measure, and it out-performs state-of-the-art self-training and augmentation-based methods trained on the same seed labels by $>13\%$.
\end{itemize}

\textbf{Reproducibility}: The source code to reproduce the experiments can be found at \url{https://github.com/xinyangz/ltrn}.

\section{Preliminaries}
\label{sec:prelim}

In this section, we first formally define the problem, and then introduce the concept of the text-rich network and its construction.

\vspace{-3mm}
\subsection{Problem Formulation}
Structure-rich text categorization takes a set of documents $\cD = \{d_1, ..., d_N\}$ accompanied with metadata $\cM = \{m_1, ..., m_N\}$, a set of labels $\cL = \{l_1, ..., l_{|\cL|} \}$, and aims to assigns a label $l_i$ for each document $d_i \in \cD$.
Metadata refers to the complementary attributes coming with textual documents in the data collection, such as writers and publishers of books, brands and manufacturers of e-commerce products, high-quality phrases mined from the text, and label surface names that are mentioned in the corpus.
In the minimally-supervised setting, the user gives a few (in our framework, no more than 3) labeled \textit{seed documents} per class $\cD_L=\bigcup_{l \in \cL}\cD_l \subset \cD$.
Distinct from the conventional supervised setting, the number of seed documents is small $|\cD_L| \ll |\cD|$; both labeled $\cD_{L}$ and unlabeled documents $\cD - \cD_{L}$ are adopted to learn the categorization model.

\vspace{-3mm}
\subsection{Text-Rich Network}
\label{sec:text-rich}
To facilitate minimally-supervsed text categorization, we build a \textit{text-rich network} from the corpus.
It provides a holistic view of the raw text documents and various document metadata.

Formally, a text-rich network $G=(D, V, E, \phi, \psi, \omega)$ has a set of document raw text $D$, a set of nodes $V$, and a set of edges $E$. 
Nodes $V=(V_T, V_A)$ are divided into textual nodes $V_T$ and auxiliary nodes $V_A$, where textual nodes have associated textual descriptions and auxiliary nodes serve as bridges connecting textual nodes together.
$\phi$ and $\psi$ are type mappings that maps each node $v$ to its type $\phi(v)$ and each edge $e$ to its relation $\psi(e)$.
Mapping $\omega: V_T \mapsto D$ is a one-to-one content mapping that maps each textual node to its raw text content.





Take the e-books dataset in our experiments as an example (Figure~\ref{fig:text-rich}).
The raw text documents are book titles and descriptions.
The textual nodes in the network are the books.
The auxiliary nodes are authors, publishers, high-quality phrases, and label surface names.


Specifically, the metadata nodes used in our experiments can be grouped into the following categories.

\smallsection{Document Attributes as Nodes}
We cast all distinct document attribute values into nodes in the text-rich network.
For instance, all authors and publishers in an e-book collection; all brands and manufacturers in an e-commerce product collection.
The edge weights from each document to its attribute nodes are set to 1.

\smallsection{High-Quality Phrases as Nodes}
We use AutoPhrase~\cite{shang2018automated} to mine high-quality phrases from the textual documents and present them as nodes in the network.
The edge weight from a phrase to a document is the TF-IDF score of that phrase in the document.
People have used other strategies for text-rich network construction, such as using all words in the corpus vocabulary as nodes~\cite{yao2019textgcn, mekala2020meta}.
We include only high-quality phrases into our network because they will not introduce an overwhelmingly large number of nodes and work well in practice.

\smallsection{Label Surface Name as Nodes}
Label surface name is another important source of information when supervision is scarce.
For example, on e-commerce platforms, merchants often put a product's category into its title; on movie review sites, the movie introduction may mention a movie's genre.
While the occurrence of a label surface name in a document does not necessarily imply the document belongs to that category, it represents a relation between the document and the label in many cases.
As opposed to distant supervised methods, we do not leverage label name matching for hard or soft labeling; instead, we add an edge between a document and a matched label surface name and set the edge weight to the label number of occurrences in the document text.


\section{Our Framework}
\label{sec:method}

In this section, we describe our proposed framework and introduce its key components in detail.

\subsection{Overview}

    
    The proposed \our is a pseudo labeling framework where the model assigns pseudo labels to unlabeled examples in each of its iterations, and gradually improves the performance through iterations.
    The general workflow of \our is shown in Figure.~\ref{fig:framework}.
    After constructing a text-rich network from the corpus, \our relies on two major components to conduct categorization and pseudo labeling.
    The first component is a text analysis module.
    We employ the BERT model~\cite{devlin2019bert} in this module and use it for both document classification and embedding.
    The second component is a network learning module.
    We propose a novel GNN model for class-discriminative, scalable learning.
    The model takes network structure as well as node features as input, and generates prediction on textual nodes (corresponding to raw text documents) in the network.
    Distinct from most existing works, our framework models both raw text and network structure, and we learn both modules in parallel to let them mutually enhance each other.
    
    Putting two modules together, we introduce a joint training method with two techniques -- co-training and feature sharing.
    In each iteration of the framework, the text module and the network module are trained separately on both user given seed documents and pseudo labeled documents.
    Once both modules are updated, we generate two sets of pseudo labels from the two modules, and pool them together to update the shared pseudo label set.
    Thereafter, each module could benefit from its own confident predictions as well as the confident predictions from the other module by co-training on the shared pseudo label set.
    At the end of each iteration, we generate up-to-date document embedding using the BERT model, and share the embedding with the GNN model as features of the textual nodes.
    When the model is fully trained, we use the BERT model for new document categorization.



\subsection{Network Learning Module}
The network learning module aims to build a class-discriminative, scalable machine learning model for effective propagation of categorical information on the text-rich network.
We propose a novel GNN architecture for the \textit{semantic-aware} propagation of minimal supervision.
The model takes both the network structure and the node features as input, and can be learned end-to-end.
Once trained, the network module assigns pseudo labels to the unlabeled nodes in the network, and the most confident pseudo labels will be selected to improve the text analysis counterpart of the framework.



We highlight two design requirements of the GNN architecture: class-discriminativeness and scalability.
For class-discriminativeness, the model must be able to handle the text-rich network constructed from a non-label discriminative way.
The network exhibits weak homophily, meaning that an edge in the network does not necessarily imply the two nodes connected by it belongs to the same category.
Popular GNN architectures such as Graph Convolutional Networks~\cite{kipf2017gcn} do not parameterize the feature aggregation process, making them suboptimal in networks with weak homophily, thus do not satisfy our needs.
Our model must be able to identify the more ``important'' neighbors in a node's mixed local neighborhood, i.e., which of the neighbors are more label-indicative in terms of offering meaningful semantics for category prediction on the current node.
For scalability, real-word corpora, especially unlabeled corpora collected automatically, are usually gigantic.
We cannot afford full batch GNN models, as the network wouldn't fit into (GPU) memory.

To meet the aforementioned requirements, we propose a novel GNN architecture 
that is capable of learning class-discriminativeness on the text-rich network, in a mini-batch fashion.
The model uses personalized PageRank~\cite{haveliwala2002ppr} (PPR) for neighborhood sampling, and an attention mechanism for feature aggregation.

The overall architecture of the GNN model involves two stages, feature transformation and neighborhood aggregation.
For each node, the model applies the following operations:

\begin{align}
    \vh_i &= f_{\mathrm{trans}}(\vx_i), \quad \vz_i = g_{\mathrm{agg}}(\vh_i, \vH, G)
\end{align}

where $f_{\mathrm{trans}}$ is a transformation function on node features, and $g_{\mathrm{agg}}$ is an aggregation function which aggregates hidden representation from the node's neighborhood and combines it with the target node's representation.
In our model, $f_{\mathrm{trans}}$ is a multi-layer feed-forward neural network, and we will introduce $g_{\mathrm{agg}}$ in the following subsections.

\subsubsection{PPR-based Neighborhood Sampling}
The first step of feature aggregation in a GNN is to define a node's neighborhood and select proper neighbors for aggregation.
Most GNN models~\cite{kipf2017gcn,hamilton2017graphsage,velickovic2018gat} aggregates from a node's first-hop neighbors, and extends to multi-hop neighbors through multiple aggregation layers.
As the size of the neighborhood grows exponentially with the number of hops, to scale up to large datasets, we sample a fixed number of nodes from the neighborhood.

Recent works have shown that personalized PageRank (PPR) is very effective in both graph neural networks ~\cite{klicpera2019appnp, bojchevski2020pprgo} and in propagating weak supervision ~\cite{mekala2020meta}.
We, therefore, adopt PPR for neighborhood sampling in GNN.

Personalized PageRank~\cite{haveliwala2002ppr} can be derived from a random walk with restart process on the network.
It takes the network structure as input, and computes a ranking score $P_{i,j}$ from each node $j$ on the network to a target node $i$.
The larger the $P_{i,j}$, the more ``similar'' node $j$ is to node $i$.
Let $\vP \in \R^{N \times N}$ be the personalized PageRank matrix of the graph, where each row of the matrix $\vP_{i, :}$ corresponds to a PPR vector to a target node $i$.
$\vP$ is defined as the solution to the following equation:

\begin{equation}
    \vP = \beta \hat{\vA} \vP + (1-\beta) \vI
\end{equation}

where $\beta$ is the reset probability for PPR and $\hat{\vA}$ is the normalized adjacency matrix.
PPR is well studied and can be approximated efficiently, even for very large networks~\cite{tong2006fast,anderson2006push}.
Similar to ~\cite{bojchevski2020pprgo}, we use a push iteration method~\cite{anderson2006push} to efficiently compute PPR scores.

After solving the PPR matrix $\vP$, we define the PPR sampled neighborhood of node $i$ as its top-K PPR neighbors:

\begin{equation}
    \cN(i) = \argmax_{V' \subset V_T, |V'| = K} \sum_{v_j \in V'} P_{i, j}
\end{equation}

Note that we only select textual nodes as top PPR neighbors because only textual nodes have meaningful text embedding features. $K$ is a small value compared to a nodes' full neighborhood size. We find $K=50$ is good enough in our experiments.

\subsubsection{Attention-Based Aggregation}
Top neighbors sampled by PPR are usually quite high quality; however, two key issues are still unresolved.
First, the PPR computation only leverages the network structure, thus is unaware of node features and semantics; second, the sampled neighbors are NOT guaranteed to be label-indicative.
Instead of computing a weighted sum of neighborhood representation using edge weights or PPR scores, we would like to parameterize the aggregation process, so that the model can learn to focus on more label-indicative nodes.

We adopt an attention-based aggregation strategy:

\begin{align}
\label{eqn:attn}
    \vz_i &= g_{\mathrm{agg}}(\vh_i, \vH, G) = \sigma \del{\sum_{j\in \cN(i)} \alpha_{i,j} P_{i, j} \vh_j}\\
    \alpha_{i,j} &= \mathrm{Sigmoid}\del{\vW_q \vh_i \cdot \vW _k\vh_j}
\end{align}

where $\vW_q$ and $\vW_k$ are learnable weights.
The idea is to first map hidden representation of the target node $\vh_i$ and the neighbor node $\vh_j$ into a shared attention space, then compute their dot product and map it to $\alpha \in \sbr{0,1}$.
The smaller the dot product is, the less relevant the neighbor is to the categorization of the current node.
The attention weight $\alpha$ is then used to adjust the personalized PageRank score of the neighbor $P_{i,j}$, before a weighted sum aggregation of all neighboring node representations.
Our attention formulation
is slightly different from that of GAT~\cite{velickovic2018gat}, as GAT does not explicitly model the dot product similarity of a pair of nodes.

\subsection{Text Analysis Module}
Our framework is compatible with any text classifier that takes raw text as input.
We adopt a pre-trained BERT~\cite{devlin2019bert} model in this work as it has been proven to be generalizable, and we have found it to work well given a small number of labeled examples.

The BERT model is a Transformer~\cite{vaswani2017attention} language model pre-trained on large, general domain corpora.
Input sentences to the BERT model are preceded with a [CLS] token and succeeded with a [SEP] token.
The model is trained using a masked language model (MLM) objective and a next sentence prediction objective.
We fine-tune the BERT model with the language model objectives on the domain corpus before using it for document categorization and embedding.

\smallsection{Document Categorization}
We append one linear layer to the model, which takes the [CLS] token embedding from the final layer of the Transformers, and produce categorical predictions on the label space.
Together with the added linear layer, the whole model is fine-tuned on labeled data, with a larger learning rate for the final layer and a smaller learning rate for the Transformer layers.

\smallsection{Document Embedding}
We adopt two strategies for document embedding generation.
The first strategy works before the model is fine-tuned on labeled data, and we use this method to generate initial embedding for textual nodes in the text-rich network.
We take the token embedding from the final Transformer layer, discard [CLS] and [SEP] token embedding, and take the average of the rest of the token embeddings as the sentence embedding.
We discard [CLS] and [SEP] token embedding because [CLS] is used for next sentence prediction in the pre-training, making it irrelevant to categorization, and both tokens have no specific meaning.
We find this strategy to work better than taking [CLS] embedding or using an average of all token embeddings as sentence representation.

The second strategy works when the model is fine-tuned on labeled data.
This method is adopted from the second to the last iteration of the framework.
We use the [CLS] embedding from the final Transformer layer as the sentence embedding.
This is because once the model is fine-tuned, [CLS] captures the categorical semantic of the sentence.

\subsection{Joint Training of Text and Network Modules}
Now we have introduced our network and text learning modules, we are ready to put them together through a joint training framework.
Algorithm~\ref{alg:train} describes the joint training procedure.
We employ two techniques for training with weak supervision: co-training and feature sharing.

\smallsection{Co-Training}
Since both of the modules in our framework are learning models, we leverage co-training to let them mutually enhance each other.
The basic idea is to take confident predictions from one model and add them to the other model's training set.
As two modules have different inductive biases, they usually generate different predictions, making the pseudo labeled training set diverse.

In each of the training iteration, we run both models on the unlabeled document set, and take the most confident predictions from each model.
We set a confidence threshold to filter out low confident predictions, and take top $M$ predicted documents from each category as pseudo labeled documents for that category.
The choice of these parameters is described in Section~\ref{sec:exp_config}.
We pull those predictions together to form a pseudo labeled training set, and join it with the given seed documents.
Both models will be trained on the new training set in the next iteration.
This process is repeated until the pseudo label set stays the same, or until a maximum iteration is reached.
We found the model to work well within 5 iterations.

\smallsection{Feature Sharing}
The performance of our GNN module relies on the quality of node features.
It cannot process raw text, and has to take vectorized features as input.
In our framework, we generate document node features from the text module.
Initially, the features are generated by an unsupervised BERT model.
As the training progress, the text module gets improved, and could produce higher quality embedding for the documents.
Therefore, we share the most up-to-date features from the text module with the GNN module.
We have found that such a feature sharing mechanism is beneficial for the GNN model's performance.

\begin{algorithm}[t]
    \SetAlgoLined
    \LinesNumbered
    \KwIn{Text-rich network $G$ (including all Documents $D$), seed (labeled) documents $D_S$.}
    Initialize training document set $T = D_S$\;
    Initialize text module $f$ and network module $g$\;
    Initialize document embedding $X \leftarrow \mathrm{embed}(f, D)$\;
    \Repeat{max\_iter or $T$ doesn't change}{
        $f \leftarrow$ train\_text\_module($T$) \;
        $g \leftarrow \mathrm{train\_network\_module}(T, X, G)$ \;
        $T_1 \leftarrow \mathrm{take\_confident\_prediction}(f, D - D_S)$\;
        $T_2 \leftarrow \mathrm{take\_confident\_prediction}(g, D - D_S)$\;
        $X \leftarrow \mathrm{embed}(f, D)$\;
        $T \leftarrow S \cup T_1 \cup T_2$\;
    }
    \caption{Joint Training of \our}
    \label{alg:train}
\end{algorithm}


\section{Experiments}
    \label{sec:exp}
    
    In this section we answer the following research questions with experiments on real-world datasets.
    \begin{itemize}[leftmargin=*,nosep]
        \item How does our model perform against state-of-the-art methods with minimal supervision?
        \item How do the text and network modules of our framework mutually enhance each other through joint training?
        \item Is the GNN architecture design of our network module helpful?
        \item How does the number of labeled seed documents impact the model performance?
    \end{itemize}
    

\subsection{Datasets}
    
    We conduct our experiments on two real-world data collections: Amazon products and Goodread Books.
    The Books dataset is adapted from ~\cite{mekala2020meta,wan2018books}, while the Amazon dataset is a larger scale dataset with many more categories collected by us.
    The dataset statistics are shown in Table~\ref{tbl:dataset}.
    For our main experiments, we provide 3 seed documents per category as supervision.
    The seed documents are randomly chosen from the entire labeled training set.
    We will study the effects of seed document set in Section~\ref{sec:seed}.
    
     \begin{table}[t]
        \centering
        \small
        \setlength\tabcolsep{2pt}
        \caption{Dataset Statistics.}\label{tbl:dataset}
        \vspace{-3mm}
        \begin{tabular}{ccccccc}
            \toprule
                 & \#doc & \#category & \#train-labeled & \#train-total &\#dev & \#test\\
            \midrule
            \textbf{Amazon} & 104,787 & 683 & 2,049 & 41,914 & 10,479 & 52,394\\
            \midrule
            \textbf{Books} & 33,594 & 8 & 24 & 21,163 & 2,354 & 10,079  \\
            \bottomrule
        \end{tabular}
        \vspace{-3mm}
    \end{table}
    
    \begin{itemize}[leftmargin=*,nosep]
        \item \textbf{Amazon.}
            We collected $\sim$100K products from Amazon.com spanning 683 product categories.
            The document attributes include the product brand and product manufacturer (e.g., ``Seventh Generation'' and ``Unilever Group'' as the brand and manufacturer of a bleach product).
            The textual description of each product has three parts:
            (1) a title, which is a concise summary of the product;
            (2) a bullet point list, highlighting product features;
            (3) a long description, introducing the product in more detail.
            We concatenate all three parts together in the above order because it follows the same ordering a typical customer views a product.
            If the full text is too long for the model, this concatenation strategy ensures that the model sees the most important part of the product text.
            
            We labeled the products through a crowdsourcing platform.
            In this way, we ensure that the labels are high-quality.
            We do not rely on category information in the Amazon catalog, as that information might be machine generated and may be incorrect.
            
            This dataset has far more categories than the datasets from prior works~\cite{miyato2017vat,meng2018westclass, mekala2020meta, li2019learn-self}.
            Many of the categories are close to each other, e.g. ``screw driver'' vs. ``wrench'', ``paper towel holder'' vs. ``hanging rod''.
            The dataset will help us benchmark different methods under a fine-grained setting.
            
        \item \textbf{Books~\cite{wan2018books,mekala2020meta}}.
            The Books dataset contains book descriptions from a popular online book review website named Goodreads\footnote{\url{https://www.goodreads.com/}}.
            We follow ~\cite{mekala2020meta} and select a subset of books from 8 popular genres\footnote{Categories in Books dataset: (1) children, (2) comics, (3) fantasy, (4) history \& biography, (5) crime, mystery thriller, (6) poetry, (7) romance, (8) young adult.} from the original dataset~\cite{wan2018books}, resulting in a dataset with 33,594 books.
            The task is to predict the genre of each book.
            The document attributes include the author and publisher of each book.
            The dataset has 22,145 distinct authors and 5,186 publishers.
            The textual part of each book contains its title and description.
            Compared to the Amazon dataset, we have a much smaller number of categories.
            Keeping the number of labeled seed documents the same, the dataset has only 24 labeled documents in total.
    \end{itemize}
    
    We use an inductive setting throughout our experiments.
    The models have access to the user-given seed documents and the unlabeled documents in the training set, while the models at training do not see the test set documents.
    This setting is different from the ones in ~\cite{meng2018westclass,zhang2020metacat,mekala2020meta} as they use a transductive setting.

\subsection{Compared Methods}
    We evaluate \our against a variety of baseline methods.
    Towards handling weak supervision, we include representative works using (1) data augmentation, (2) pseudo labeling, and (3) combined methods.
    By input data structure, we compare against (1) text-based methods, (2) graph-based methods, and (3) combined methods.
    
    \begin{itemize}[leftmargin=*,nosep]
        \item \textbf{BERT-supervised.}
            BERT~\cite{devlin2019bert} is the de facto pre-trained language model for natural language understanding.
            We fine-tune a BERT model using all the documents from the training set as our reference supervised model.
            
        \item \textbf{BERT-seed.}
            We train a BERT model using only seed documents.
            This sets our text-based baseline without specific techniques to handle weak supervision.
            
        \item \textbf{BERT-self-train.}
            We train a BERT model using the seed documents, as well as self-training on unlabeled documents.
            This method resembles our text-based pseudo labeling baseline.
            
        \item \textbf{VAT.}
            Virtual adversarial training~\cite{miyato2017vat} perturbs the training examples in the feature space towards the worst direction, and train the classification model on the perturbed ``pseudo examples'' to improve model robustness.
        
        \item \textbf{EDA}
            ~\cite{inui2019eda} is a text data augmentation method with four simple operations: synonym replacement, random insertion, random swap, and random deletion.
            We apply EDA to our seed document set and train a BERT model with EDA.
            
        \item \textbf{UDA}
            ~\cite{xie2019uda} is the state-of-the-art data augmentation technique for deep neural network training.
            It performs back translation and TF-IDF word replacement to augment both the labeled and unlabeled data.
            After that, it trains a BERT classifier with a supervised cross-entropy loss and an unsupervised consistency loss.
            
            
        \item \textbf{WeSTClass}
            ~\cite{meng2018westclass} combines data augmentation and pseudo labeling by generating pseudo documents and employing self-training.
            It is a pre-BERT classification model.
            We use its CNN variant as it performs the best over all the model variants.
        
        \item \textbf{PPRGo-seed.}
            PPRGo~\cite{bojchevski2020pprgo} is a state-of-the-art scalable GNN model with personalized PageRank based neighborhood aggregation. We train it on seed document only as our network-based baseline.
        
        \item \textbf{PPRGo-self-train.}
            We train PPRGo with seed documents and self-training on unlabeled documents. This method resembles our network-based pseudo labeling baseline.
        
        \item \textbf{Text-GCN}
            ~\cite{yao2019textgcn} constructs a corpus-level network of words and documents with word-document edges based on TF-IDF and word-word edges based on PMI.
            It then applies the Graph Convolutional Network~\cite{kipf2017gcn} for document categorization.
            
        \item \textbf{Text-ING}
            ~\cite{zhang2020texting} converts each document into a network, where nodes are words in the documents and edges are word occurrences based on a sliding window. After converting documents to networks, a GNN is applied for classification. 
            
        \item \textbf{CANE}
            ~\cite{tu2017cane} is a textual network embedding method which applies a convolutional neural network for text representation, and maximizes edge probabilities with a combination of structure-based and text-based objectives. 
        
        \item \textbf{MetaCat}
            ~\cite{zhang2020metacat} learns embedding for words, documents, labels, and metadata to categorize text documents with minimal supervision. It combines network embedding with a CNN text classifier, and employs data augmentation.
            
    \end{itemize}
    
    We denote our method as \textbf{\our}.
    We use Micro-F1 (i.e., accuracy) and Macro-F1 as our evaluation metrics.

\subsection{Model Configuration}
    \label{sec:exp_config}
    
    We append metadata as a string to the main text for all text-based baselines to ensure a fair comparison.
    For example, if a product from the Amazon dataset has a brand ``Clorox'', then we append ``[BRAND] Clorox'' to the product's description.
    For PPRGo models, we use unsupervised BERT embedding as node feature.
    
    For all methods using the BERT model, we use BERT-base architecture with pre-trained weights from the original authors and adapted by HuggingFace Transformers library\footnote{\url{https://huggingface.co/transformers/pretrained_models.html}}.
    We then fine-tune it using masked language model objective on domain-specific corpus with a $10^{-5}$ learning rate.
    For the PPRGo model as well as our GNN submodule, we set the number of layers to 2, and the hidden dimension to 64.
    We set max sequence length to 128 for all models, and ensure that the metadata string appended to the end of the text is not truncated.
    During BERT model fine-tuning, we set the learning rate to $10^{-4}$ for Transformer layers, and $10^{-2}$ for the classification layer.
    
    For the CANE model, since it operates on homogeneous textual networks, we convert our heterogeneous text-rich networks to document-document networks.
    Denote $D_M$ as the document to metadata edge matrix, we obtain document to document edges and their weights by $D_M D_M^{T}$.
    
    As for our model, we set $K=50$ for top $K$ PPR neighbor sampling in GNN.
    We set the confidence filtering threshold to be $0.9$ for co-training, and take top $50$ and $500$ documents for each category in Amazon and Books dataset, respectively.
    We use different top prediction numbers because the number of unlabeled documents per category is larger in the Books dataset.
    One can also use a top percentage of confident predictions for co-training.
    The same setting is applied to all self-training methods.
    We set the maximum number of iteration in our framework to be 5.
    
\vspace{-3mm}
\subsection{Main Results}

    We present our main experiment results in Table~\ref{tbl:main_results}.
    Our proposed \our model consistently outperforms competing baseline methods on both datasets.
    We also note the following key observations throughout our experiments.
    
    \begin{itemize}[leftmargin=*,nosep]
        \item On the Amazon dataset, text-based methods have an advantage over their network-based counter parts.
        While on the Books dataset, network-based methods achieve higher performances.
        The joint training framework of our model takes advantage of both sources of information, thus achieving superior performance than the baselines.
        \item BERT representations generalize well on both datasets even when the supervision is scarce.
        The models that do not use BERT representation (VAT, WeSTClass, MetaCat, Text-GCN) could suffer given very few training labels.
        The significant performance gain of our model over BERT baselines demonstrates that the network module enhanced BERT performance.
        \item Comparing Text-ING to other network-based methods, it shows inferior performance.
        We attribute this observation to the minimally supervised setting, which limits the training vocabulary and the Text-ING model's generalization power.
        CANE also suffers in our experiments, as our text-rich network has weak edge homophily. CANE relies on maximizing all edge probabilities, thus may not be ideal for this application scenario.
        \item Data augmentation methods have great success on the Amazon dataset, EDA being our best performing baseline.
        However, they struggle with the Books dataset.
        Comparing UDA to EDA, we attribute its inferior performance to the unsatisfactory result of back-translation augmentation on the e-commerce product data and the poor convergence of the model on the Books dataset.
        While we do not incorporate data augmentation methods into our framework, \our effectively learns from unlabeled data.
        Note that EDA data augmentation can also be added to our framework to further boost the model performance.
    \end{itemize}
    
    Comparing our model to the supervised reference model, the performance difference is around 8\% on the Books dataset and less than 2\% on the Amazon dataset, which is quite impressive given that the supervised model has $800\times$ more labeled data on the Books dataset and $20\times$ more on the Amazon dataset.
    
    \begin{table}[t]
        \center
        \caption{Evaluation Results on Two Datasets. Methods are grouped into \textit{supervised}, \textit{text-based}, \textit{network-based}, and \textit{ours}. Underline marks the best score within each group.}
        \vspace{-3mm}
        \label{tbl:main_results}
        \small
        \scalebox{0.95}{
            \begin{tabular}{l cc cc}
                \toprule
                                 & \multicolumn{2}{c}{\textbf{Amazon}} & \multicolumn{2}{c}{\textbf{Books}} \\
                \cmidrule{2-5}
                \textbf{Methods} & Micro-F$_1$ & Macro-F$_1$ & Micro-F$_1$ & Macro-F$_1$ \\
                \midrule
                BERT-supervised~\cite{devlin2019bert} & \textbf{0.938} & \textbf{0.921} & \textbf{0.853} & \textbf{0.855} \\
                \midrule
                BERT-seed~\cite{devlin2019bert} &  0.679 & 0.660 & 0.448 & 0.439 \\
                BERT-self-train & 0.751  & 0.733 & \underline{0.599} & \underline{0.611} \\
                VAT~\cite{miyato2017vat} & 0.336 & 0.321 & 0.305 & 0.265 \\
                EDA~\cite{inui2019eda} & \underline{0.793} & \underline{0.781} & 0.472 & 0.487  \\
                UDA~\cite{xie2019uda} & 0.748  & 0.711  & 0.434  & 0.414    \\
                WeSTClass~\cite{meng2018westclass} & 0.337 & 0.315  & 0.387 & 0.404   \\
                \midrule
                CANE~\cite{tu2017cane} & 0.218 & 0.136 & 0.283 & 0.262 \\
                MetaCat~\cite{zhang2020metacat} & 0.389 & 0.374 & 0.419  & 0.437 \\
                Text-GCN~\cite{yao2019textgcn} & 0.684 & \underline{0.674} & 0.505 & 0.494 \\
                Text-ING~\cite{zhang2020texting} & 0.456 & 0.438 & 0.483 & 0.485 \\
                PPRGo-seed~\cite{bojchevski2020pprgo} & 0.647 & 0.601 & 0.604  & 0.612 \\
                PPRGo-self-train & \underline{0.687} & 0.659 & \underline{0.691}  & \underline{0.697} \\
                \midrule
                \our & \textbf{0.921} & \textbf{0.905} & \textbf{0.774} & \textbf{0.778} \\
                \bottomrule
            \end{tabular}
        }
        \vspace{-3mm}
    \end{table}

\subsection{Case Studies and Ablation Studies}
\label{sec:case_ablation}
\definecolor{ao}{rgb}{0.0, 0.5, 0.0}
    \begin{table*}[t]
        \center
        \caption{Different High-Confidence Predictions from Network and Text Modules.
        }
        \vspace{-3mm}
        \label{tbl:case_predictions}
        \small
        \begin{tabular}{p{2.2in}p{1.2in}ccc}
            \toprule
            Document & Category & BERT Model Prediction & GNN Model Prediction  \\
            \midrule
            BlueDot Trading Dictionary Secret Book Hidden Safe with Key Lock,... & Safe & \makecell{(\textcolor{red}{Writing Paper}, 0.031) \\ (\textcolor{red}{Self Stick Note}, 0.027)} & (\textcolor{ao}{Safe}, 0.917) \\
            \midrule
            Puoyis Toddler Kids Swim Life Vest, Girls and Boys Swim Float Jacket... & Water Flotation Device & \makecell{(\textcolor{red}{Dress}, 0.044) \\ (\textcolor{red}{Incontinence Protector}, 0.041)} & (\textcolor{ao}{Water Flotation Device}, 0.987) \\
            \midrule
            \midrule
            Logitech 910-000253 VX Nano Cordless Laser Mouse & Input Mouse &  (\textcolor{ao}{Input Mouse}, 0.907) & \makecell{(\textcolor{ao}{Input Mouse}, 0.154) \\ (\textcolor{red}{Keyboard}, 0.010)} \\
            \midrule
            PDP 048-082-NA-CM05 Stealth Series Wired Controller for Xbox One, Xbox One X & Game Controller & (\textcolor{ao}{Game Controller}, 0.905) & \makecell{(\textcolor{red}{Game Console}, 0.149) \\ (\textcolor{red}{Keyboard}, 0.125)} \\
            \bottomrule
        \end{tabular}
    \end{table*}
    
    \begin{figure*}
        \centering
        \includegraphics[width=\linewidth]{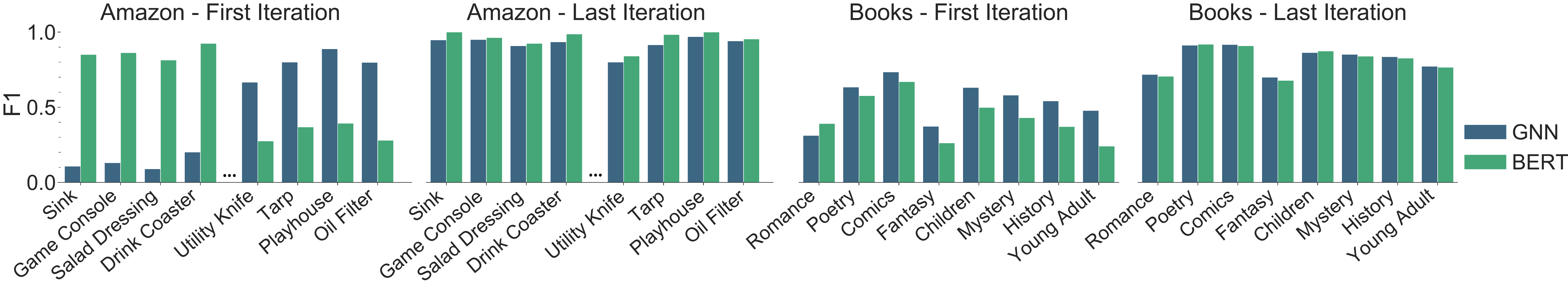}
        \vspace{-8mm}
        \caption{Per Category Performance Change for Text and Network Modules Over Iterations.}
        \vspace{-3mm}
        \label{fig:case_performance}
    \end{figure*}

    In this section, we present case studies and ablation studies to scrutinize our proposed framework.
    Specifically, we aim to answer the following research questions: (1) How do the text and network modules mutually enhance each other through joint training? (2) Is our GNN design helpful for text-rich network modeling?

    \smallsection{Different High-Confidence Predictions from Network and Text Modules}
    In the first case study, we show that the GNN model from the network module and the BERT model from the text module have different inductive biases, thus making different high confident predictions.
    We take high confident predictions $T_1$, $T_2$ from both models, as shown in Line 7,8 in Algorithm~\ref{alg:train}.
    Next, among the most confident predictions $T_2$ from GNN, we select those examples with \textbf{least} BERT confidence.
    Similarly, we also select examples where BERT is most confident, but GNN is least confident.
    The chosen examples show where one model does very well while the other model does poorly.
    We conduct this case study on the Amazon dataset.
    The examples are chosen from the very first iteration of the framework.
    
    The results are shown in Table~\ref{tbl:case_predictions}.
    In the first two rows, we show two products where GNN performs well, but BERT does not.
    The first product is a hidden safe disguised as a book, and the second product is a life vest.
    These two products are quite challenging judging from textual descriptions only. The first product has wordings similar to a book, and the second product mentions words like ``kids'', ``jacket'' that are commonly seen in apparel products.
    From the BERT model prediction, we observe that it confuses the first product with paper, notes, and confuses the second product with dresses, incontinence protectors.
    The GNN model, on the other hand, successfully classifies them due to the information from the text-rich network structure, such as their brand and manufacturers.
    
    The third and fourth row contains two examples where BERT does well, but GNN does not.
    The GNN cannot do well in those cases because products's neighbors come from a different category, and may mislead the GNN model.
    Starting from the brands ``Logitech'', ``PDP'', we find other peripheral devices of different categories; similarly, the key phrases, e.g., ``Xbox One'', ``Xbox One X'', connects the game controller product to game consoles through the network structure.
    Indeed, from the GNN predictions, we see that it confuses these products with ``keyboard'' and ``game console''.
    
    The examples in the case study show that two modules have different strengths in the beginning iteration, and that there is an opportunity for one model to learn from the other model.
    
    \smallsection{Per Category Performance Change Over Iterations}
    After observing that the network module and the text module learn to make different predictions in the first iteration,
    we are interested in whether both modules can mutually enhance each other through joint training.
    Therefore, we present a second case study on the per-category performance change of BERT and GNN models.
    
    As shown in Figure~\ref{fig:case_performance}, we present BERT and GNN model performance on different categories.
    We show results at the end of the first and the last framework iteration.
    In each subfigure, the categories are ordered according to the performance difference of BERT and GNN.
    From left to right, $\mathrm{F1_{BERT}} - \mathrm{F1_{GNN}}$ goes from largest to smallest.
    On the Amazon dataset, we select eight categories where two models have the most performance difference at the first iteration.
    
    We can see that, in the beginning, the two models have different strengths in different categories.
    While in the end, the performance gaps of the two models are significantly narrowed, and the overall performance on these categories are much higher.
    Judging from the results, we can conclude that both modules are enhancing each other through joint training.
    
    \smallsection{Effect of GNN Neighbor Sampling and Attention}
    We give two cases on the Amazon dataset to show how our proposed GNN architecture can select label-discriminative neighbors from mixed neighborhoods.
    As shown in Table~\ref{tbl:case_neighbor}, we compare direct neighbors of two products to the GNN re-ranked neighbors.
    The ordering of direct neighbors is according to the edge weight in the network, while the ordering of GNN neighbors is by $\alpha_{i,j} P_{i,j}$, according to Eqn. (\ref{eqn:attn}) where the attention values are from our fully trained GNN model.
    As we can see, in the two given cases, direct neighbors of the products are quite mixed, but our GNN model can rank meaningful neighbors higher after fully trained.
    
\begin{table}[t]
    \center
    \caption{Performance of \our Model Ablations.
    }
    \label{tbl:ablation}
    \small
    \vspace{-3mm}
    \scalebox{0.95}{
    \begin{tabular}{l cc cc}
        \toprule
                         & \multicolumn{2}{c}{\textbf{Amazon}} & \multicolumn{2}{c}{\textbf{Books}} \\
        \cmidrule{2-5}
        \textbf{Methods} & Micro-F$_1$ & Macro-F$_1$ & Micro-F$_1$ & Macro-F$_1$ \\
        \midrule
        \our & 0.921 & 0.905 & 0.774 & 0.778 \\
        \our (GraphSAGE) & 0.898 & 0.871 & 0.756 & 0.759 \\
        \our (no-feat-share) & 0.872 & 0.850 & 0.748 & 0.760 \\
        \bottomrule
    \end{tabular}
    }
\end{table}

    \smallsection{Ablation Study}
    We present two ablations of our model in Table~\ref{tbl:ablation} in order to verify the effectiveness of our proposed GNN architecture design and the feature sharing mechanism in joint training.
    
    In the second row of Table~\ref{tbl:ablation}, we replace our proposed GNN with a GraphSAGE model~\cite{hamilton2017graphsage}.
    Unlike our model, GraphSAGE randomly samples node neighbors, and aggregates their features with a neural network model.
    We use the GCN variant of the GraphSAGE model, which does a weighted combination of neighbor features using normalized graph Laplacian as weights.
    On both datasets, we see a performance drop.
    We attribute the inferior performance of the model ablation to the lower quality pseudo labels generated by GraphSAGE compared to our proposed GNN architecture.
    
    In the last row of Table~\ref{tbl:ablation}, we remove the feature sharing mechanism and use the fixed unsupervised BERT embedding as GNN node features throughout the joint training.
    We can see a performance drop compared to \our.
    Without the help of updated BERT embedding, the GNN model produces slightly worse pseudo labels in later iterations and affects the overall categorization performance.
    
    \subsection{Seed Document Set Analysis}
    \label{sec:seed}
    \begin{figure}[t!]
        \centering
        \includegraphics[width=\linewidth]{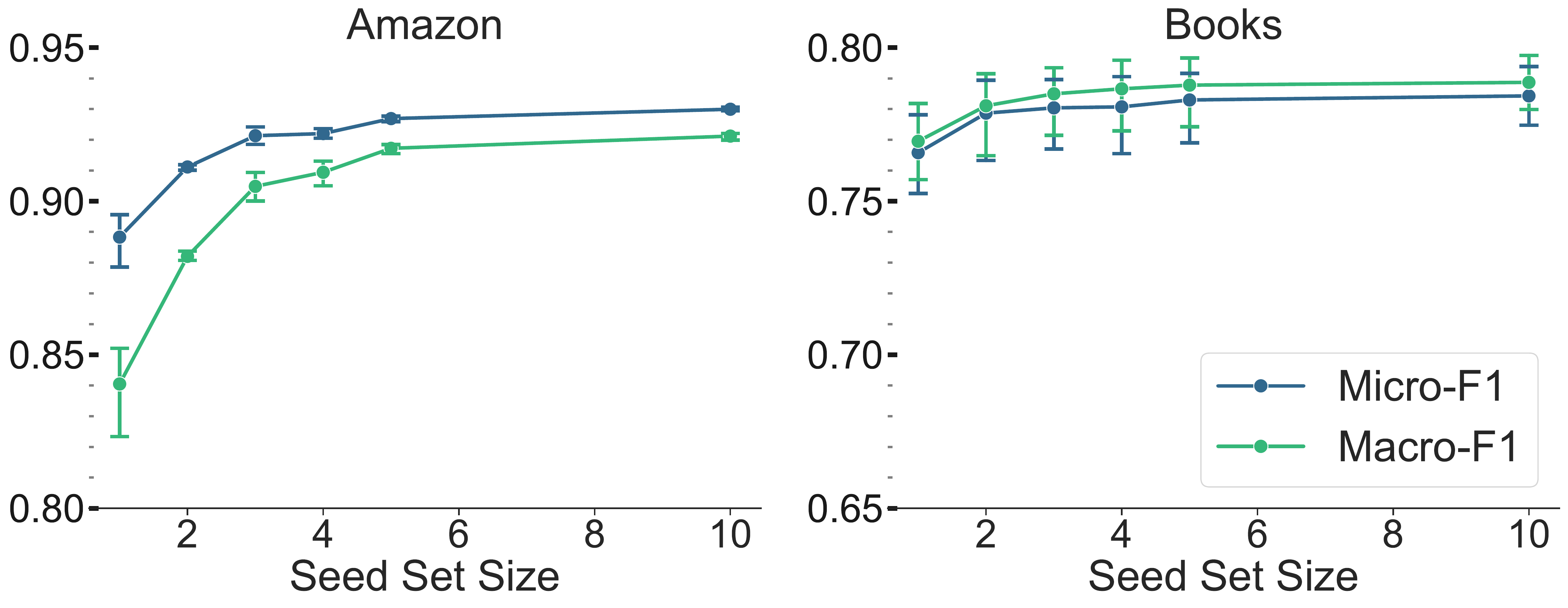}
        \vspace{-3mm}
        \caption{\our performance w.r.t \#seed documents.}
        \vspace{-3mm}
        \label{fig:seed_set_size}
    \end{figure}
    In this section, we aim to answer two research questions: (1) How does \our's performance change with respect to the number of seed documents? (2)How does the selection of labeled seed document impact model performance?
    In Figure~\ref{fig:seed_set_size}, we present different runs of our \our model using different number of seed documents.
    We run 4 separate experiments for each seed set size.
    We use a different random selection of seed documents in each run.

    As shown in Figure~\ref{fig:seed_set_size}, the performance of the \our model generally increases when the number of labeled seed documents increases.
    The trend is more evident on the Amazon dataset than the Books dataset.
    It is also worth noting that when given only 1 seed example per category, the model already performs quite well on both datasets.
    Although on the Amazon dataset, we see a bigger gap between the Micro-F1 score and the Macro-F1 score when the number of seed documents is small, suggesting that the model's performance has a larger difference between categories in the most extreme minimal supervision setting.
    
    We present the model's performance variance through error bars in the line plots.
    In general, there are two major sources of model variance: the selection of seed document as input and the randomness in the model itself.
    The result shown in Figure~\ref{fig:seed_set_size} is a combination of both sources.
    Nevertheless, we can see that the model performance variance is less than $\sim$3\% on two datasets.
    The model has a smaller variance on the Amazon dataset, and the variance gap becomes smaller as the number of seed documents increases.
    While on the Books dataset, the model has a larger variance, and the variance stays consistent for up to 10 seed documents per category.




\section{Related Work}
\label{sec:rel}

Recent studies have shown increasing interest in training a good categorization model with little human effort.
\textit{Semi-supervised learning} aims to leverage both labeled and unlabeled data to boost a model's performance.
\textit{Weakly-supervised learning} incorporates other forms of supervision, such as giving seed words for each category~\cite{meng2018westclass,mekala2020meta} or using only label surface names for classification~\cite{meng2018westclass, meng2020lotclass}.
\textit{Few-shot learning} focuses on model generalization, where the model is trained with a diverse set of categories and tested on new categories with a few examples~\cite{feifei2006oneshot, vinyals2016matching, snell2017prototypical}.
Our minimally supervised setting resembles semi-supervised learning in extreme cases, where the number of labeled examples for each category is no more than 3.

\begin{table}[t]
    \center
    \caption{Ranked Neighbors after Network Learning vs. Direct Neighbors. $\times$ indicates neighbors whose category do not match with the given item. }
    \vspace{-3mm}
    \label{tbl:case_neighbor}
    \small
    \begin{tabular}{ll}
    \toprule
    \multicolumn{1}{c}{\textbf{GNN Ranked Neighbor}} & \multicolumn{1}{c}{\textbf{Direct Neighbor}} \\
    \toprule
    \multicolumn{2}{c}{\textbf{Item}: Bico Red \& Blue Christmas Gnome Salt \& Pepper Shaker Set...} \\
    \midrule
    Bico Airtight Salt \& Pepr. Shaker & Bico Airtight Salt \& Pepr. Shaker  \\
    \midrule
    WL Disney Salt \& Pepper Shaker... & \textcolor{red}{$\times$} Garlic Press, S.S. Mincer \\
    \midrule
    WL Ceramic Salt \& Pepper Shaker... & \textcolor{red}{$\times$} YF Chopper Crusher Machine  \\
    \toprule
    \multicolumn{2}{c}{\textbf{Item}: Fortem Ratchet Tie Down Straps, 4x 15ft...}\\
    \midrule
    Rhino Ratchet Straps 4pk & \textcolor{red}{$\times$} Bovke Sci. Calculator Case\\
    \midrule
    Rhino Ratchet Straps Heavy Duty & \textcolor{red}{$\times$} Houseables Dog Tunnel \\
    \midrule
    Autofonder 4 pk ratchet tie down & \textcolor{red}{$\times$} Zomei Portable Tripod\\
    \bottomrule
    \end{tabular}
    \vspace{-3mm}
\end{table}


\smallsection{Pseudo Labeling Methods}
Combining labeled and unlabeled data for model training, pseudo labeling methods try to assign ``pseudo labels'' for unlabeled examples and incorporate them into model training.
Self-training~\cite{mihalcea2004self, yarowsky1995self} is arguably the most common strategy for pseudo label training.
In each iteration of the algorithm, high confident predictions from the model on the unlabeled dataset are added to the training set.
Recent work improves self-training by assigning weights to pseudo labeled examples~\cite{li2019learn-self} with confident learning.
Other extensions of self-training have explored using multiple models for pseudo label generation, e.g., co-training~\cite{blum1998cotrain}, tri-training~\cite{zhou2005Tritrain}, democratic co-learning~\cite{zhou2004democratic}.
Besides techniques that work on the input space, methods that incorporate pseudo examples by perturbation on the feature space have also achieved success~\cite{miyato2017vat}.
Although our method adopts the pseudo labeling framework, we jointly train two models on both text and network data, as opposed to most prior works that only employ text data.

\smallsection{Data Augmentation Methods}
This line of work augments the existing data examples to enlarge the training set.
Recently, people have found data augmentation to work well with deep neural networks~\cite{xie2019uda,inui2019eda,kobayashi2018contextual,qiu2020easyaug}.
Wei et al.~\cite{inui2019eda} perform four simple operations: synonym replacement, random insertion, random swap, and random deletion.
The augmented dataset consistently boosted the performance of deep learning-based text classifiers.
Xie et al.~\cite{xie2019uda} use back translation and TF-IDF word replacement to augment both labeled and unlabeled data. They leverage a joint loss function to tie both parts together.
While our method does not use data augmentation, it is compatible with popular text augmentation techniques and could incorporate them into training.

Another related thread to our framework is graph semi-supervised learning.
Graph neural networks (GNNs) bring deep learning to graph-structured data by message passing on graphs~\cite{kipf2017gcn, gilmer2017message}.
Most GNN models assume fixed vectorized node features, while in our work, raw text coming with the nodes plays a predominant role in categorization.
Pioneer studies bringing text and network together typically use a text-centric model to maximize edge probabilities in a graph~\cite{tu2017cane, shen2018align} or use a network-centric model to learn node representations~\cite{zhang2020metacat} or use a fixed random walk process to propagate label on the graph~\cite{shang2020nettaxo, mekala2020meta}.
In this paper, we bring text and graph learning together for minimal supervision.




\section{Conclusions \& Future Work}
\label{sec:con}

In this paper, we proposed a minimally supervised text categorization model by learning on text-rich networks.
We leverage a BERT model for raw text understanding and proposed a novel GNN model to model label-discriminative signal in the text-rich network structure.
We jointly train the BERT model with the GNN model with co-training and feature sharing.
With only a few user given seed examples, the model shows competitive performance even when compared to a supervised model.

In the future, we would like to explore models that can capture heterogeneous type information in the text-rich network.
We would also like to incorporate label type taxonomy into our framework.

\smallsection{Acknowledgments}
This work was supported in part by US DARPA KAIROS Program No. FA8750-19-2-1004 and SocialSim Program No.  W911NF-17-C-0099, National Science Foundation IIS-19-56151, IIS-17-41317, and IIS 17-04532, and the Molecule Maker Lab Institute: An AI Research Institutes program supported by NSF under Award No. 2019897. 
Any opinions, findings, and conclusions or recommendations expressed in this document are those of the author(s) and should not be interpreted as the views, either expressed or implied, of DARPA or the U.S. Government


\balance
\bibliographystyle{abbrv}
\bibliography{main}

\end{document}